%% file: unsupervised_awe_icassp2019.tex
\documentclass{article}

\usepackage{spconf}

\usepackage{graphicx}
\graphicspath{{fig/}}

\usepackage{booktabs}
\usepackage{tabularx}
\usepackage{multirow}
\newcommand{\mytable}{
    \centering
    \renewcommand{\arraystretch}{1.1}
    }
\newcolumntype{C}{>{\centering\arraybackslash}X}
\newcolumntype{L}{>{\raggedright\arraybackslash}X}
\newcolumntype{R}{>{\raggedleft\arraybackslash}X}
\newcolumntype{P}[1]{>{\raggedright\arraybackslash}p{#1}}

\usepackage{amsmath}
\usepackage{amssymb}
\renewcommand{\vec}[1]{\boldsymbol{{#1}}}

\usepackage{microtype}
\usepackage{url}
\usepackage{xcolor}
\newcommand{\system}[1]{{\small \textsc{#1}}}
\newcommand\blfootnote[1]{\begingroup
                          \renewcommand\thefootnote{}\footnote{#1}
                          \addtocounter{footnote}{-1}
                          \endgroup}

\usepackage{cite}
\let\oldbibliography\thebibliography
\renewcommand{\thebibliography}[1]{\oldbibliography{#1}
                                   \setlength{\itemsep}{-0.34mm}
                                   \vspace*{-0mm}}

\title{Truly unsupervised acoustic word embeddings using weak top-down constraints in encoder-decoder models}
\name{Herman Kamper}
\address{E\&E Engineering, Stellenbosch University, South Africa \\
         {\small \tt kamperh@sun.ac.za}}

\usepackage[prependcaption,textsize=scriptsize]{todonotes}
\begin{document}


\maketitle

\input{abstract}
\input{introduction}

\input{model}
\input{experiments}
\input{conclusion}

\newpage
\ninept
\bibliography{icassp2019}

\end{document}

%% file: abstract.tex
\begin{abstract}
We investigate unsupervised models that can map a variable-duration speech segment to a fixed-dimensional representation. In settings where unlabelled speech is the only available resource, such \textit{acoustic word embeddings} can form the basis for ``zero-resource'' speech search, discovery and indexing systems. Most existing unsupervised embedding methods still use some supervision, such as word or phoneme boundaries. Here we propose the encoder-decoder correspondence autoencoder (\system{EncDec-CAE}), which, instead of true word segments, uses automatically discovered segments: an unsupervised term discovery system finds pairs of words of the same unknown type, and the \system{EncDec-CAE} is trained to reconstruct one word given the other as input. We compare it to a standard encoder-decoder autoencoder (AE), a variational AE with a prior over its latent embedding, and downsampling. \system{EncDec-CAE} outperforms its closest competitor by 29\% relative in average precision on two languages in a word discrimination task.
\end{abstract}

%% file: introduction.tex
\section{Introduction}

Current speech recognition models require large amounts of annotated resources. 
For many languages, the transcription of speech audio remains a major obstacle~\cite{adda+etal_sltu16}. 
This~has~prompted research into \textit{zero-resource speech processing}, which aims to develop methods that 
can discover linguistic structure and representations directly from unlabelled speech~\cite{jansen+etal_icassp13,versteegh+etal_sltu16,dunbar+etal_asru17}.
This problem also has strong links with early language acquisition, since infants learn language without explicit hard supervision~\cite{rasanen_speechcom12,elsner+shain_emnlp17}.

Several zero-resource tasks have been tackled. 
In unsupervised term discovery (UTD), 
the aim is to find recurring word- or phrase-like patterns in an unlabelled speech collection~\cite{park+glass_taslp08}. 
In query-by-example search, the goal is to identify utterances containing instances of a given spoken query~\cite{hazen+etal_asru09,zhang+glass_asru09,levin+etal_icassp15,wang+etal_icassp18}.
Full-coverage segmentation and clustering aims to tokenise an entire speech set into word-like units~\cite{lee+etal_tacl15,rasanen+etal_interspeech15,kamper+etal_asru17,elsner+shain_emnlp17,bhati+etal_interspeech17}.
In all these tasks, a system needs to compare speech segments of variable length.
Dynamic time warping (DTW) is traditionally used, but is computationally expensive and has known~\mbox{limitations~\cite{rabiner+etal_tassp78}}.

Recent studies have therefore explored an alignment-free methodology where
an arbitrary-length speech segment is embedded in a fixed-dimensional space such that segments of the same word type have similar embeddings~\cite{maas+etal_icmlwrl12,levin+etal_asru13,chung+etal_interspeech16,audhkhasi+etal_stsp17,settle+etal_interspeech17,chung+etal_nips18,chung+glass_interspeech18,holzenberger+etal_interspeech18,chen+etal_slt18}. 
Segments can then be compared by simply calculating a distance in this \textit{acoustic word embedding} space.
Several unsupervised methods have been proposed, ranging from using distances to a fixed reference set~\cite{levin+etal_asru13}, to unsupervised auto-encoding encoder-decoder recurrent neural networks~\cite{chung+etal_interspeech16,audhkhasi+etal_stsp17}.
Downsampling, where a fixed number of equidistant frames are used to represent a segment, has proven to be an effective 
and strong baseline~\cite{rasanen+etal_interspeech15,holzenberger+etal_interspeech18}.
Many of the more advanced unsupervised models, however, 
still use some form of supervision, normally in the form of known word boundaries~\cite{levin+etal_asru13,chung+etal_interspeech16,chung+glass_interspeech18}.

We propose a new neural model which is truly unsupervised, assuming no labelled speech data or word boundary information.
We use a UTD system---itself unsupervised---to find pairs of word-like segments predicted to be of the same unknown type.
Each pair is presented 
to an autoencoder-like encoder-decoder network: one word in the pair is used as the input and the other as the target output.
The latent representation between the encoder and decoder is used as acoustic embedding.
We call this model the encoder-decoder correspondence autoencoder (\system{EncDec-CAE}). 
The idea is that the model should learn an intermediate representation that is invariant to properties not common to the two segments (e.g.\ speaker, channel), while capturing aspects that are (e.g.\ word identity).


We compare this model to downsampling~\cite{levin+etal_asru13} and an encoder-decoder autoencoder~\cite{chung+etal_interspeech16} trained on random segments.
We also propose another model: an encoder-decoder variational autoencoder, which incorporates a prior over its latent embedding.
We show that the 
\system{EncDec-CAE} outperforms the other models in a  word discrimination task on English and Xitsonga data.
In Xitsonga, it even outperforms a DTW approach which uses full alignment to discriminate words. 

%% file: model.tex
\section{Acoustic Word Embedding Methods}

We consider three unsupervised neural models in this work, all using an 
encoder-decoder recurrent neural network (RNN) architecture~\cite{cho+etal_emnlp14,sperduti+starita_tnn97}.
The first model was proposed as an acoustic embedding method in~\cite{chung+etal_interspeech16}, while the other two are~new.


\subsection{Encoder-decoder autoencoder} 
\label{sec:ae}


An encoder-decoder RNN consists of an encoder RNN, which reads in an input sequence while sequentially updating its internal hidden state, and a decoder RNN, which produces an output sequence conditioned on the final output of the encoder~\cite{cho+etal_emnlp14}.
Chung et al.~\cite{chung+etal_interspeech16} 
trained an encoder-decoder as an autoencoder (\system{EncDec-AE}) on unlabelled speech segments,
using the input of the network as the target, as shown in Figure~\ref{fig:encdec}(a).
The final fixed-dimensional output vector $\vec{z}$ from the encoder (red in Figure~\ref{fig:encdec}) gives an acoustic word embedding. 


More formally, an input segment $X = \vec{x}_1, \vec{x}_2, \ldots, \vec{x}_T$ consists of a sequence of frame-level acoustic feature vectors $\vec{x}_t \in \mathbb{R}^D$ 
(e.g.\ MFCCs).
The loss for a single training example is 
$\ell(X) = 
\sum_{t = 1}^T \left|\left|\vec{x}_t - \vec{f}_t(X)\right|\right|^2$, with
$\vec{f}_t(X)$ 
the $t^\textrm{th}$ decoder output conditioned on the latent embedding $\vec{z}$, which itself is produced as the output of the encoder when it is fed with~$X$.

In our implementation of the \system{EncDec-AE} we use a transformation of the final hidden state of the encoder RNN to produce the embedding $\vec{z} \in \mathbb{R}^M$, as also in~\cite{audhkhasi+etal_stsp17}. 
We use gated recurrent units~\cite{chung+etal_arxiv14} as the RNN cell type in all the models here.
This worked better than LSTMs in validations experiments.
Most importantly, instead of using true word segments~\cite{chung+etal_interspeech16}, we train the \system{EncDec-AE} on random speech segments. 

\subsection{Encoder-decoder variational autoencoder} 

A variational autoencoder (VAE) is a generative neural model
which uses a variational approximation for inference and training~\cite{kingma+welling_arxiv13}.
A complete treatment of VAEs is outside the scope of this work, so we refer readers to~\cite{kingma+welling_arxiv13,doersch_arxiv16}.
Here we only describe how we develop the encoder-decoder VAE (\system{EncDec-VAE}), where the decoder can be seen as a generative sequence model conditioned on a latent representation $\vec{z}$, which we use as acoustic embedding.
We show that by choosing specific distributions, the model can be interpreted as a standard \system{EncDec-AE} with a prior over its latent embedding.

\begin{figure}[t]
    \centering
    \includegraphics[width=0.85\linewidth]{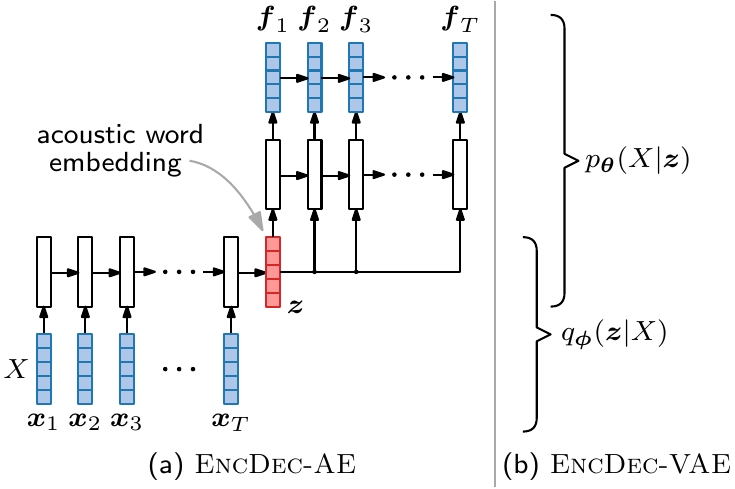}
    \vspace*{-10pt}
    \caption{
    The \system{EncDec-AE} is trained to predict its input (a speech segment) from the latent acoustic word embedding $\vec{z}$.
    The \system{EncDec-VAE} is a generative model which incorporates a prior over the latent acoustic embedding $\vec{z}$.}
    \label{fig:encdec}
    \vspace*{-7pt}
\end{figure}

For the \system{EncDec-VAE}, it is useful to think of the encoder and decoder as separate 
networks with parameters $\vec{\phi}$ and $\vec{\theta}$, respectively.
The encoder density is denoted as $q_{\vec{\phi}}(\vec{z} | X)$ and the decoder as $p_{\vec{\theta}}(X|\vec{z})$, as in Figure~\ref{fig:encdec}(b).
For a single training sequence $X$, the \system{EncDec-VAE} maximises a lower bound for
\vspace*{-7.5pt}
\begin{equation}
    \log p_{\vec{\theta}}(X) = \log 
    \prod_{t = 1}^T p_{\vec{\theta}} (\vec{x}_t | \vec{x}_{< t}) 
    = \sum_{t = 1}^T \log p_{\vec{\theta}} (\vec{x}_t | \vec{x}_{< t})
    \label{eq:loss_single} \vspace*{-7.5pt}
\end{equation}
By approximating the intractable posterior $p_{\vec{\theta}}(\vec{z}|X)$ using the encoder $q_{\vec{\phi}}(\vec{z} | X)$, each of the terms in the above summation can be bounded, as is done in the standard VAE~\cite{kingma+welling_arxiv13}:
\vspace*{-2.5pt}
\begin{align}
    \log p_{\vec{\theta}} (\vec{x}_t | \vec{x}_{< t}) 
    &\geq \mathbb{E}_{q_{\vec{\phi}}(\vec{z}|X)} \left[ \log p_{\vec{\theta}}(\vec{x}_t| \vec{x}_{< t}, \vec{z}) \right] \nonumber \\
    &\qquad - D_{\textrm{KL}} \left( q_{\vec{\phi}}(\vec{z}|X) || p(\vec{z}) \right) \label{eq:elbo}
    = J_t(X) 
    \\[-17.5pt] \nonumber
\end{align}
with $D_{\textrm{KL}}$ denoting the Kullback-Leibler divergence and $J_t$ the evidence lower bound 
for frame $\vec{x}_t$.
The expectation in~\eqref{eq:elbo} is often approximated as
$\mathbb{E}_{q_{\vec{\phi}}(\vec{z}|X)} \left[ \log p_{\theta}(\vec{x}_t|\vec{x}_{<t}, \vec{z}) \right] \approx \log p_{\theta}(\vec{x}_t|\vec{x}_{<t}, \vec{z}')$, 
using the single Monte Carlo sample $\vec{z}' \sim q_{\vec{\phi}}(\vec{z}|X)$. The reparametrisation trick is used to backpropagate through this sampling operation. 
Choosing a spherical Gaussian distribution for the decoder output $p_{\vec{\theta}}(\vec{x}_t|\vec{x}_{<t}, \vec{z})$ with mean $\vec{f}_t(\vec{z}')$ and fixed covariance $\sigma^2\mathbf{I}$, this approximation term reduces to
$\log p_{\theta}(\vec{x}_t|\vec{x}_{<t}, \vec{z}') = c - \frac{1}{2\sigma^2} \left|\left| \vec{x}_t - \vec{f}_t(\vec{z}') \right|\right|^2$,
with $c$ a density normalisation constant 
and $\vec{f}_t (\vec{z}')$ the $t^{\textrm{th}}$ output of the decoder network given the sampled latent variable $\vec{z}'$.
The KL-term in~\eqref{eq:elbo} also has a closed-form solution 
when assuming a standard Gaussian prior $p(\vec{z}) = \mathcal{N}(\vec{z}; \vec{0}, \mathbf{I})$ and a diagonal-covariance Gaussian distribution for the encoder $q_{\vec{\phi}}(\vec{z}|X) = \mathcal{N} ( \vec{z};\vec{\mu}_{\vec{\phi}}(X), \mathrm{diag}(\vec{\sigma}^2_{\vec{\phi}}(X)) )$; see~\cite[Appx~B]{kingma+welling_arxiv13}. 
These mean $\vec{\mu}_{\vec{\phi}}(X)$ and covariance $\vec{\sigma}^2_{\vec{\phi}} (X)$ vectors 
are produced as outputs of the encoder given input sequence $X$.

Taking together the above equations, 
the \system{EncDec-VAE}'s loss for a single training sequence $X$ is thus
\vspace*{-7.5pt}
\begin{align}
    \ell(X)
    &= -\sum_{t = 1}^T J_t(X) \nonumber \\[-3pt] 
    &= \sum_{t = 1}^T \left[ \frac{1}{2\sigma^2} \left|\left| \vec{x}_t - \vec{f}_t(\vec{z}') \right|\right|^2 + D_{\textrm{KL}} \left( q_{\vec{\phi}}(\vec{z}|X) || p(\vec{z}) \right) \right]  \nonumber \\[-22pt] \nonumber
\end{align}
This is a combination of a reconstruction term (as for the standard \system{EncDec-AE}) and a 
KL-regularisation term encouraging embedding $\vec{z}$ given input $X$ to be close to the prior distribution $p(\vec{z})$.
The $\sigma$ hyperparameter 
controls the relative weight of reconstruction vs.\ regularisation.
We use $\sigma = {10}^{-5}$ (optimised in validation), so high emphasis is placed on reconstruction. 

One motivation for the \system{EncDec-VAE} is that in the \system{EncDec-AE} we have empirically observed that embeddings for words of different types often take on very similar values across embedding dimensions.
By specifying a prior, we obtain a handle on the spread or concentration of embedding values. 






\begin{figure}[t]
    \centering
    \includegraphics[width=0.95\linewidth]{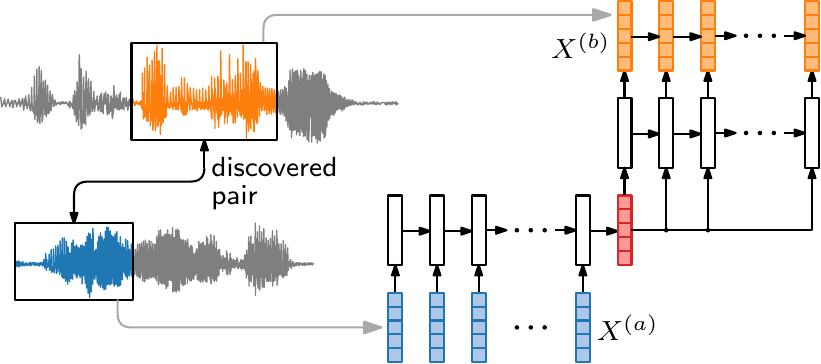}
    \vspace*{-8pt}
    \caption{An unsupervised term discovery system is used to obtain input-output speech segments for the encoder-decoder correspondence autoencoder (\system{EncDec-CAE}).}
    \label{fig:cae}
    \vspace*{-7pt}
\end{figure}

\subsection{Encoder-decoder correspondence autoencoder} 

While the \system{EncDec-AE} uses the same speech segments as input and output, the encoder-decoder correspondence autoencoder (\system{EncDec-CAE}) uses weak top-down constraints in the form of discovered word pairs to have input and output sequences from different instances of the same predicted word type.
This is illustrated in Figure~\ref{fig:cae}.
An unsupervised term discovery~(UTD) system is applied to an unlabelled speech collection, discovering pairs of word segments predicted to be of the same type.
These are then presented as input-output pairs to the \system{EncDec-CAE}.
Since UTD is itself unsupervised, the overall approach is unsupervised.

For \system{EncDec-CAE}, a single training item consists of a pair of sequences $(X^{(a)}, X^{(b)})$. 
Each segment consists of a unique sequence of frame-level vectors: {$X^{(a)} = \vec{x}_1^{(a)}, \ldots, \vec{x}_{T_a}^{(a)}$} and {$X^{(b)} = \vec{x}_1^{(b)}, \ldots, \vec{x}_{T_b}^{(b)}$}.
The loss for a single training pair is 
\mbox{$\ell(X^{(a)}, X^{(b)})
    = \sum_{t = 1}^{T_b} ||\vec{x}^{(b)}_t - \vec{f}_t(X^{(a)})||^2$},
where $X^{(a)}$ is the input and $X^{(b)}$ the target output. 
We found that it is crucial to first pretrain the \system{EncDec-CAE} as a standard AE using the loss in \S\ref{sec:ae}, before switching to the loss here.

The correspondence idea is not new---its application in the encoder-decoder architecture is.
In~\cite{renshaw+etal_interspeech15} it was used to improve frame-level rather than segmental speech representations. 
Earlier~\cite{silberer+lapata_acl14} and more recent work~\cite{hashimoto+etal_nips17} in domains outside of speech have also used 
neighbours in the input feature space as model targets. 
Term discovery systems were also used to provide training segments for a speech model in~\cite{chung+etal_nips18}.
Although their systems are unsupervised, the intrinsic quality of embeddings was not their direct focus, as is also the case in~\cite{wang+etal_icassp18,audhkhasi+etal_stsp17}.
Another study where ground truth segments are not used is~\cite{holzenberger+etal_interspeech18}, although true phoneme boundaries are~assumed.


%% file: experiments.tex
\section{Experiments}

\subsection{Experimental setup and evaluation}


We perform experiments on two languages, using the same data as in previous zero-resource studies~\cite{versteegh+etal_sltu16}.
English training, validation and test sets are obtained from the Buckeye corpus~\cite{pitt+etal_speechcom05}, each with around 6~hours of speech.
For Xitsonga we use a 2.5~hour portion of the NCHLT corpus~\cite{devries+etal_speechcom14}.
We use word pairs discovered using the UTD system of~\cite{jansen+vandurme_asru11}. 
On the English training data, this system discovers around 14k unique pairs; on the Xitsonga data, it discovers around 6k pairs.
For comparison, we also extract a similarly sized set of ground truth word segments from forced alignments of the English training data.
All speech audio is parametrised as static Mel-frequency cepstral coefficients~(MFCCs), i.e.\ $D = 13$.

Ultimately we want to use acoustic embeddings downstream in zero-resource speech applications.
But here we want to measure intrinsic quality without being tied to a particular system architecture.
We therefore use a word discrimination task designed 
for this purpose~\cite{carlin+etal_icassp11}.
In the \textit{same-different task}, we are given a pair of acoustic segments, each a true word, and we must decide whether the segments are examples of the same or different words.
To do this using an embedding method, a set of words in the test data (we use around 5k tokens in both languages) are embedded using the specific approach.
For every word pair in this set, the cosine distance between their embeddings is calculated.
Two words can then be classified as being of the same or different type based on some threshold, and a precision-recall curve is obtained by varying
the threshold.
The area under this curve is used as final evaluation metric, referred to as the average precision (AP).

As a baseline embedding method, we use downsampling by keeping 10 equally-spaced MFCC vectors from a segment with appropriate interpolation, giving a 130-dimensional embedding.
This has proven a strong baseline in other work~\cite{holzenberger+etal_interspeech18}.
The same-different task can also be approached by using DTW alignment between test segments, where the alignment cost of the full sequences are used as a score for word discrimination.


All neural models are implemented in TensorFlow. 
For all models we use an embedding dimensionality of $M = \textrm{130}$, to be directly comparable to the downsampling baseline.
More importantly, although other studies consider higher-dimensional settings, downstream systems such as~\cite{kamper+etal_asru17} are constrained to embedding sizes of this order.
Neural network architectures were optimised on the English validation data.
Our final \system{EncDec} models all have 3 encoder and 3 decoder unidirectional RNN layers, each with 400-unit hidden representations. We also considered bidirectional RNNs, but this did not give improvements.
Pairs are presented to the \system{EncDec-CAE} in both input-output directions.
Models are trained using Adam optimisation~\cite{kingma+ba_iclr15} with a learning rate of 0.001.
On the English data, early stopping is used, but for Xitsonga validation data is not available so the models are simply trained for 100~epochs.
We run all models with five different initialisations and report average performance and standard deviations.

\begin{table}[!t]
    \mytable
    \caption{Word discrimination performance on test data. UTD was used to provide training segments for the \system{EncDec} models. 
    }
    \vspace{5pt}
    \renewcommand{\arraystretch}{1.1}
    \begingroup
    \small
    \begin{tabularx}{1.0\linewidth}{@{}lCC@{}}
        \toprule
        & \multicolumn{2}{c}{Average precision (\%)} \\
        \cmidrule(l){2-3}
        Model & English & Xitsonga  \\
        \midrule
        \system{EncDec-AE} & 24.8 $\pm$ 0.50 & 14.2 $\pm$ 0.30 \\
        \system{EncDec-VAE} & 25.0 $\pm$ 0.20 & 11.4 $\pm$ 0.40 \\ 
        \system{EncDec-CAE} & \textbf{{32.2}} $\pm$ 0.01 & \textbf{{32.0}} $\pm$ 0.02 \\
        \addlinespace
        Downsampling & 21.7 & 13.6  \\
        DTW alignment & 35.9 & 28.1 \\
        \bottomrule
    \end{tabularx}
    \endgroup
    \label{tbl:results}
    \vspace*{-7pt}
\end{table}

\subsection{Results}

Table~\ref{tbl:results} shows AP performance on test data for all the neural embedding approaches, the downsampling baseline, and DTW alignment.
All \system{EncDec} models here are trained on UTD segments, yielding truly unsupervised results.
The dimensionalities of all the embeddings are also identical, $M = \textrm{130}$. 

\system{EncDec-CAE} outperforms all other embedding approaches in both languages.
In English it outperforms its closest competitor (\system{EncDec-VAE}) by roughly 29\% relative in AP, while for Xitsonga it achieves more than twice the AP of \system{EncDec-AE}.
In Xitsonga, \system{EncDec-CAE} also outperforms DTW, which has access to the full uncompressed sequences.
This improvement is particularly notable since DTW is computationally more expensive:
embedding comparisons with \system{EncDec-CAE} takes about 0.5 minutes on a single CPU core while DTW takes more than 60 minutes parallelised over 20 cores.
In previous studies where embeddings were reported to outperform DTW, either ground truth word segments~\cite{levin+etal_asru13} or higher-dimensional embeddings were used~\cite{holzenberger+etal_interspeech18}.

\begin{figure}[!b]
    \vspace*{-2pt} 
    \vspace*{-7pt}
    \centering
    \includegraphics[width=0.85\linewidth]{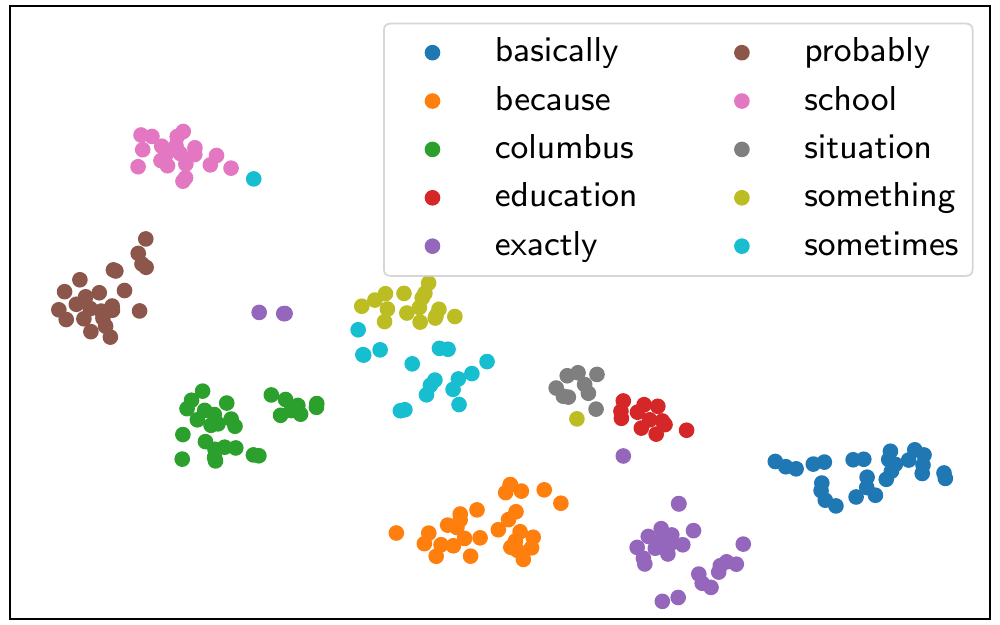}
    \vspace*{-7pt}
    \caption{t-SNE visualisations of acoustic word embeddings produced by \system{EncDec-CAE} on unseen data.}
    \label{fig:tsne}
\end{figure}

In Table~\ref{tbl:results}, all \system{EncDec} models outperform downsampling apart from the Xitsonga \system{EncDec-VAE}. 
But downsampling is simple and does not require any training.
In English, \system{EncDec-VAE} performs similar to \system{EncDec-AE}.
This is not surprising since we had to
weigh the reconstruction term in the VAE loss relatively high to obtain reasonable AP scores during validation.
The poor performance in Xitsonga is in part due to the absence of a validation set: model hyperparameters (to which \system{EncDec-VAE}  seem to be particularly sensitive) could not be tuned and early stopping was not possible. 
More work on \system{EncDec-VAE} is required to make definitive conclusions.

\vspace*{-5pt} 
\subsection{Further analysis} 
\vspace*{-3pt} 

To quantify the penalty of using discovered instead of true word segments, Table~\ref{tbl:gt_rand_utd} shows performance on the English validation data for the different \system{EncDec} models trained on different types of segments.
The oracle column indicates AP when models are trained on ground truth word segments obtained from forced alignments. 
The random column shows AP when training segments are sampled randomly from within training utterances.
\system{EncDec-AE} pays a small penalty when not using true segments, while \system{EncDev-VAE} performs best with UTD segments.
\system{EncDec-CAE} cannot be trained on random segments as it requires paired data.
When using perfect instead of discovered pairs, \system{EncDec-CAE} improves from 31.7\% to 51.1\% in AP.
This is also the best overall performance on the English validation data---better than DTW.
On the validation data here, downsampling yields performance similar to \system{EncDec-AE} and \system{EncDec-VAE}, without requiring any training segments.
Row 3 shows that pretraining the \system{EncDec-CAE} is essential as performance drops considerably without it.

Figure~\ref{fig:tsne} shows t-SNE visualisations~\cite{vandermaaten+hinton_jmlr08} of embeddings from \system{EncDec-CAE} trained on UTD segments.
Tokens of the same type are mapped to similar regions, and acoustically similar words such as `situation' and `education', or `something' and `sometimes' are located close to each other.


\begin{table}[!t]
    \mytable
    \caption{Performance on English validation data, with models  trained on ground truth, random and UTD~\mbox{segments.}
    }
    \vspace{5pt}
    \renewcommand{\arraystretch}{1.1}
    \begingroup
    \small
    \begin{tabularx}{1.0\linewidth}{@{}l@{\ \ }CCC@{}}
        \toprule
        & \multicolumn{3}{c}{Average precision (\%)} \\        
        \cmidrule{2-4}
        Model & Oracle & Random & UTD \\
        \midrule
        \system{EncDec-AE} & 26.2 & \textbf{25.5} & 25.7 \\
        \system{EncDec-VAE} & 25.8 & 25.4 & 26.0 \\
        \system{EncDec-CAE} (non-init.) & {46.9} & - & {19.0} \\
        \system{EncDec-CAE} & \textbf{{51.1}} & - & \textbf{{31.7}} \\
        \addlinespace
        Downsampling & \multicolumn{3}{c}{------ \quad 24.5 \quad  ------} \\
        DTW alignment & \multicolumn{3}{c}{------ \quad 36.8 \quad  ------} \\
        \bottomrule
    \end{tabularx}
    \endgroup
    \label{tbl:gt_rand_utd}
    \vspace*{-7pt}
\end{table}

%

%
%

%% file: conclusion.tex
\vspace*{-5pt} 
\section{Conclusion}
\vspace*{-5pt} 

We have evaluated old and new unsupervised neural acoustic word embedding methods for the truly unsupervised case where no word labels or boundaries are known.
We showed that the encoder-decoder correspondence autoencoder (\system{EncDec-CAE}) outperforms other approaches in two languages.
This model uses training pairs from a top-down unsupervised term discovery (UTD) system.
When perfect segments are used, the \system{EncDec-CAE} also performs best and consistently outperforms a full alignment-based model.
This indicates that the model could be even more effective given improvements in UTD accuracy.
We also proposed the \system{EncDec-VAE}, a variational autoencoder over segments, but this model did not give improvements.
It could, however, provide more explicit disentangled features~\cite{hsu+etal_nips17}, which we did not consider here and will investigate in future work.\blfootnote{We would like to thank Ewald van der Westhuizen for helpful {feedback}, and NVIDIA Corporation for sponsoring a Titan~Xp~GPU for this work.}

%